\documentclass{article}

 \PassOptionsToPackage{numbers, compress}{natbib}

\usepackage[main, final]{neurips_2026}  %

\usepackage[utf8]{inputenc} %
\usepackage[T1]{fontenc}    %
\usepackage{hyperref}       %
\usepackage{url}            %
\usepackage{booktabs}       %
\usepackage{tabularx}       %
\usepackage{amsfonts}       %
\usepackage{amssymb}        %
\usepackage{amsmath}        %
\usepackage{nicefrac}       %
\usepackage{microtype}      %
\usepackage{xcolor}         %
\usepackage{graphicx}       %
\usepackage{algorithm}      %
\usepackage{algorithmic}    %
\usepackage{float}          %
\usepackage{multirow}       %
\usepackage{makecell}       %
\usepackage{caption}        %
\usepackage{subcaption}     %
\usepackage{enumitem}
\usepackage[most]{tcolorbox}
\usepackage{etoc}            %
\newcommand{\std}[1]{{\tiny$\pm$#1}}  %

\newif\ifshowcomments
\showcommentsfalse
\ifshowcomments
  \newcommand{\cy}[1]{\textcolor{blue}{\textbf{[CY:]} #1}}
  \newcommand{\sk}[1]{\textcolor{red}{\textbf{[SK:]} #1}}
\else
  \newcommand{\cy}[1]{}
  \newcommand{\sk}[1]{}
\fi
\title{Curriculum Learning for Safety Alignment}

\author{%
  Sandeep Kumar \\
  Carnegie Mellon University \\
  \texttt{sandeep3@andrew.cmu.edu} \\
  \And
  Virginia Smith\footnotemark[1] \\
  Carnegie Mellon University \\
  \texttt{smithv@cmu.edu} \\
  \And
   Chhavi Yadav\thanks{Equal contribution.} \\
  Carnegie Mellon University \\
  Simons Institute, UC Berkeley \\
  \texttt{cyadav@andrew.cmu.edu} \\
}
\begin{document}

\maketitle

\begin{abstract}

Direct Preference Optimization (DPO) is a widely used approach for safety alignment that aims to reduce harmful behaviors in large language models. However, prior work shows that it can be brittle and exhibits poor out-of-distribution (OOD) generalization \citep{qi2025safety}. In this paper we investigate whether Curriculum Learning can improve the robustness of DPO-based safety alignment. We propose \textbf{Staged-Competence}, a curriculum-based framework that organizes preference data by difficulty, employs competence-based sampling, and progressively updates the reference model during training. Averaged across three model families and three training seeds, Staged-Competence reduces OOD harmful response rates by 13.6 percentage points and attack success rates across five jailbreak attacks by 16.5 percentage points, while preserving general capabilities and maintaining near-zero over-refusal. We further show that Staged-Competence: (1) matches baseline safety performance with only 75\% of the training data, demonstrating improved data efficiency and (2) yields better separation between safe and unsafe responses. Staged-Competence is agnostic to the underlying policy optimization loss and can extend to other DPO variants and alignment domains beyond safety. Additionally, we find multiple inconsistencies in standard RLHF datasets and create a cleaner combined version, named \textbf{PiKacHHU}. Our code is available at \url{https://github.com/Sandeep5500/curriculum-learning-for-safety} and the dataset at \url{https://huggingface.co/datasets/etrigan5500/PiKacHHU}.

\end{abstract}

\etocdepthtag.toc{maintag}
\section{Introduction}
Safety alignment of large language models (LLMs) seeks to ensure that models refuse harmful requests while remaining helpful on benign ones~\citep{bai2022training}.
Direct Preference Optimization (DPO)~\citep{rafailov2023direct} has emerged as a popular approach, learning from human-annotated preference pairs of safe and unsafe responses without requiring a separate reward model.
However, standard DPO has been found to be brittle to simple jailbreaking attacks~\citep{zou2023universal,qi2025safety} and fails to generalize out-of-distribution~\citep{lin2024limited,qi2023finetuning,lermen2023lora,feng2024analyzing}.

On the other hand, curriculum learning~\citep{bengio2009curriculum,hacohen2019power,soviany2022survey} has been shown to teach models more robust features by ordering examples from easy to hard, allowing the learner to build on simpler concepts before more challenging ones.
While this principle has been applied to tasks such as machine translation~\citep{platanios2019competence}, pretraining~\citep{elgaar2026curriculum,xie2023doremi}, general alignment~\citep{pattnaik2024enhancing,li2025campus}, its potential for \emph{safety} alignment remains largely unexplored. This brings us to the question:

\begin{center}
\textit{Can curriculum learning lead to more robust safety alignment?}
\end{center}

In this work, we investigate the aforementioned question and conduct the first systematic study of curriculum learning strategies for DPO-based safety alignment. Although curriculum learning is a natural tool to explore in this scenario, preference data presents a key challenge: the difficulty of a preference pair depends not just on linguistic complexity but on how well the unaligned model already distinguishes safe from unsafe behavior. To address this, we propose a difficulty score called \textit{preference alignment margin}, which orders samples by how well the model already distinguishes safe from unsafe responses. Next, we propose a curriculum training algorithm, \textit{Staged-Competence}, which gradually expands the pool of eligible examples during training through competence-based sampling and progressively updates the reference policy model between stages. 

Our experiments show the efficacy of Staged-Competence in learning robust features for safety alignment: across three model families and three training seeds, it reduces the OOD harmful response rate by 13.6 percentage points and the attack success rate across five jailbreak attacks by 16.5 percentage points without degrading general capabilities; achieves ${\sim}2.5\times$ greater reward margin separation; extends safety alignment beyond the first few tokens; matches baseline safety with 25\% less data; and scales gracefully with model size.

We also systematically ablate curriculum design choices, including ordering, within-stage sampling, and reference-model updates, showing how each affects safety robustness across three model architectures.
Additionally, we identify widespread preference pair inconsistencies in two popular safety datasets, PKU-SafeRLHF~\citep{ji2024pku} and HH-RLHF~\citep{bai2022training}, and release \href{https://huggingface.co/datasets/etrigan5500/PiKacHHU}{\textbf{PiKacHHU}}, a filtered combination of the two suitable for DPO-based safety training, alongside our code. Staged-Competence is agnostic to the underlying policy optimization loss and therefore extends beyond DPO and can also be applied to alignment domains other than safety.

\begin{figure*}[t]
\centering
\includegraphics[width=\linewidth,height=7cm]{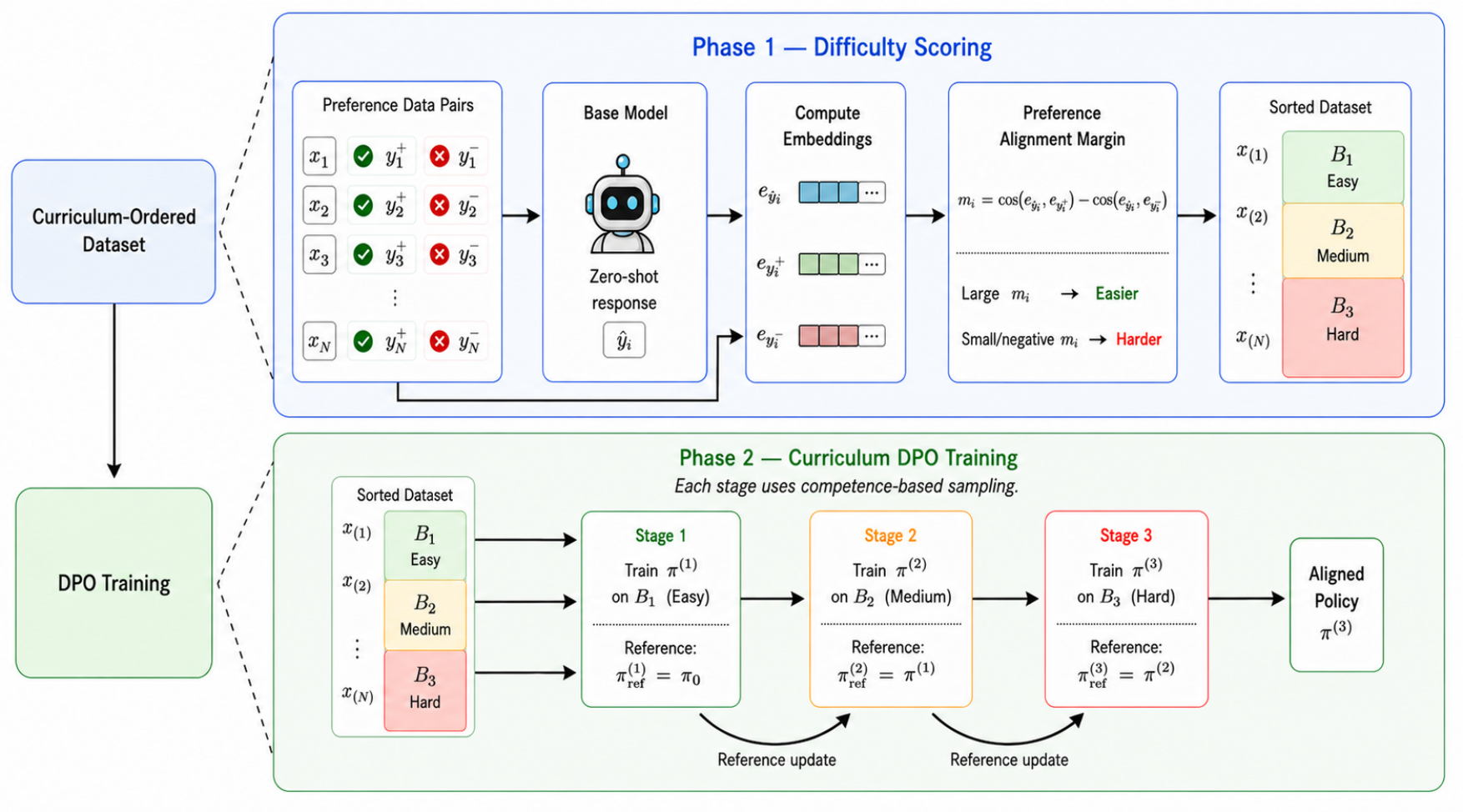}
\caption{\textit{Overview of the Staged-Competence pipeline (illustrated with $K\!=\!3$).}
\textbf{Phase~1 (Scoring):} a model-dependent preference alignment margin $m_i$ produces a global easy-to-hard ordering of preference pairs. \textbf{Phase~2 (Training):} the sorted data is split into $K$ buckets, with within-stage competence sampling and between-stage reference-model updates ($\pi_{\mathrm{ref}}^{(k+1)} = \pi^{(k)}$).}
\label{fig:method}
\end{figure*}

\section{Related Work}

\paragraph{Curriculum learning.}
\citet{bengio2009curriculum} introduced curriculum learning, showing that ordering training examples from easy to hard improves convergence over random presentation.
Subsequent work demonstrated that this ordering improves not only convergence but also final generalization in deep networks~\citep{hacohen2019power}, with a broader survey synthesizing evidence that curriculum-based ordering improves robustness and out-of-distribution generalization across machine learning domains~\citep{soviany2022survey}.
\citet{platanios2019competence} subsequently extended these ideas to neural machine translation through a \emph{competence-based} formulation, in which a growing pool of eligible examples introduces harder cases at a progressively slower rate.

\paragraph{Curriculum learning for LLMs.}
Curriculum learning has more recently been adapted to large language model training.
\citet{xie2023doremi} optimize the pretraining data mixture using a proxy model that re-weights domains over training, yielding faster convergence and stronger downstream performance.
\citet{li2025campus} introduce competence-aware curriculum scheduling for instruction tuning, dynamically adjusting example difficulty to match the model's capability.
Both target general capability rather than safety alignment, leaving curriculum-based safety alignment underexplored.

\paragraph{Curriculum methods for alignment.}
The closest prior work is Curri-DPO~\citep{pattnaik2024enhancing}, which combines curriculum learning with DPO via difficulty-stratified stages and between-stage reference-model updates; we discuss it in more detail in Section~\ref{sec:curriculum_methods}.

\paragraph{DPO safety brittleness and robustness.}
A growing body of work shows DPO-based safety alignment is fragile: \citet{qi2025safety} demonstrate that safety alignment is shallow, concentrating in the first few output tokens, and \citet{qi2023finetuning} show that even modest fine-tuning can compromise safety.
A parallel line of work targets DPO's objective: \citet{meng2024simpo} remove the reference model and length bias to stabilize training, and \citet{ethayarajh2024kto} recast DPO under prospect theory for better behavior under skewed data.
For safety specifically, \citet{zhao2025door} introduce a dual-objective DPO and \citet{kim2026safedpo} reformulate DPO with explicit safety constraints.
Our method keeps the standard DPO loss intact and is therefore compatible with these objective-level improvements.

\section{Preliminaries}

\subsection{DPO Problem Formulation}
\label{sec:problem}

Given a safety preference dataset $\mathcal{D} = \{(x_i, y_i^+, y_i^-)\}_{i=1}^{N}$ consisting of $N$ examples, where $x_i$ is an input prompt, $y_i^+$ is a safe (chosen) response, and $y_i^-$ is an unsafe (rejected) response, Direct Preference Optimization (DPO)~\citep{rafailov2023direct} trains a policy $\pi_\theta$ by minimizing the following loss:
\begin{equation}
\label{eq:dpo}
\mathcal{L}_{\text{DPO}}(\theta) = -\mathbb{E}_{(x,y^+,y^-) \sim \mathcal{D}} \left[ \log \sigma \!\left( 
\beta \left( \log \frac{\pi_\theta(y^+ \mid x)}{\pi_{\text{ref}}(y^+ \mid x)} 
- \log \frac{\pi_\theta(y^- \mid x)}{\pi_{\text{ref}}(y^- \mid x)} \right)
\right) \right].
\end{equation}
where $\pi_{\text{ref}}$ is a fixed reference policy and $\beta$ controls the strength of the deviation penalty.

In standard DPO, training examples are sampled uniformly at random at each step, and the reference policy $\pi_{\text{ref}}$ remains fixed throughout training.

\subsection{Curriculum Training Methods}
\label{sec:curriculum_methods}
\paragraph{Competence-based curriculum learning.}
A key challenge in curriculum learning is controlling the rate at which harder examples are introduced.
If new, more difficult examples are added too quickly, the learner may not have sufficient time to assimilate them before even harder ones arrive~\citep{platanios2019competence}.
The competence-based approach addresses this by maintaining a growing subset of the training data from which mini-batches are sampled: at any point during training, only examples up to a certain difficulty threshold are eligible, and this threshold expands gradually according to a schedule function.

Formally, each example is assigned a normalized difficulty $d_i = (\mathrm{rank}(i) - 1)/(N - 1)$ based on its rank in the sorted training set, where $\mathrm{rank}(i) \in \{1, \dots, N\}$ orders examples from easiest~(1) to hardest.
At training step~$t$ out of $T$ total steps, a competence function $c(t)$ determines the difficulty threshold, and only examples with $d_i \leq c(t)$ are included in the eligible pool for mini-batch sampling.
\citet{platanios2019competence} propose the square-root schedule $c(t) = \sqrt{(1 - c_0^2)\,t/T + c_0^2}$, where $c_0$ is an initial competence constant. This form ensures that harder examples are introduced at a decreasing rate, giving the model time to consolidate each wave before more are added.

This approach was originally developed for machine translation with RNNs and early Transformers; we adapt it to modern LLMs and safety alignment via DPO.

\paragraph{Curri-DPO.}
In standard DPO and the competence-based approach above, the reference model $\pi_{\text{ref}}$ remains fixed throughout training.
\citet{pattnaik2024enhancing} propose instead to partition the training data into $K$ difficulty-stratified buckets and run training in $K$ stages, updating the reference model between stages so that each stage's policy can focus on assimilating the current difficulty bucket rather than re-learning what was already acquired earlier: $\pi_{\text{ref}}^{(k+1)} = \pi^{(k)}.$

The original work sets $K\!=\!3$ buckets and operates on a training dataset where each prompt is paired with four candidate responses.
The difficulty for the curriculum ordering is defined \emph{locally} within each prompt and not across the whole dataset: the four responses are ranked $R_1$ (best) to $R_4$ (worst) by an external judge (e.g., GPT-4 or humans), and three preference pairs of increasing difficulty are formed -- $(R_1, R_4)$ easy, $(R_1, R_3)$ medium, $(R_1, R_2)$ hard -- where difficulty corresponds to the quality gap between the two responses. For each prompt, its three pairs are then placed in their corresponding buckets.
Because this local scheme provides no way to compare difficulty across prompts -- an ``easy'' pair from one prompt may be substantially harder than a ``hard'' pair from another -- examples within each bucket are simply randomly shuffled during their corresponding training stage.
Curri-DPO was originally developed for general helpfulness alignment; we build on its staged reference update mechanism and extend it to safety alignment with a global, model-dependent difficulty ordering that places every preference pair on a single comparable axis. %

\section{Methodology}
\label{sec:methodology}

Curriculum learning generally involves two components: (1) a difficulty scoring phase that defines a meaningful ordering of the training data, and (2) a training algorithm that determines how the curriculum is imposed during learning. We describe both in the context of DPO-based safety alignment and present \textbf{Staged-Competence}, a new curriculum learning framework for  safety training.

\subsection{Phase 1: Difficulty Scoring}
\label{sec:scoring}

Given a safety preference dataset $\mathcal{D} = \{(x_i, y_i^+, y_i^-)\}_{i=1}^N$, our goal is to construct a curriculum that orders training examples by their relative difficulty for the model. Instead of relying on static heuristics, we define difficulty in a \emph{model-dependent} manner, reflecting how well the current (unaligned) base model distinguishes safe from unsafe responses. This ensures that the curriculum is tailored to the particular model in question and its specific biases/behavior.

At a high level, we measure the difficulty of an example by passing the input prompt through the unaligned model, obtaining its zero-shot response, and comparing that response to the provided safe and unsafe responses. Intuitively, samples are easier if the model already produces outputs closer to the safe response, and harder if its outputs are closer to the unsafe alternative.

To operationalize this, for each prompt $x_i$, we generate a zero-shot response $\hat{y}_i$ from the base model and compute embeddings for $\hat{y}_i$, $y_i^+$, and $y_i^-$. We then define a \emph{preference alignment margin}:
\begin{equation}
\label{eq:margin}
m_i = \cos(e_{\hat{y}_i}, e_{y_i^+}) - \cos(e_{\hat{y}_i}, e_{y_i^-}),
\end{equation}
where $e(\cdot)$ denotes normalized sentence embeddings. A large positive margin indicates that the model’s output is already aligned with the safe response, while a small or negative margin indicates misalignment and thus higher difficulty.

We sort the training set in descending order of margin $m_i$ (easy-to-hard) to construct the curriculum used for DPO training.
Unlike the prompt-local scoring used in Curri-DPO, this margin provides a \emph{global} difficulty ordering across the entire dataset, allowing any two preference pairs to be compared regardless of their source prompt.

\subsection{Phase~2: Curriculum Training}

Among existing curriculum-based methods, Curri-DPO~\citep{pattnaik2024enhancing} demonstrates the value of curriculum learning for DPO-based tasks but we find that it suffers from a key limitation for effective safety alignment: each stage of Curri-DPO reverts to standard DPO with random shuffling, leaving the curriculum ordering unutilized at the within-stage level. To fill this gap, we draw on Sqrt-Competence~\citep{platanios2019competence}, whose core mechanism -- a growing pool of samples that adds harder examples at a progressively slower rate -- is a natural fit for incorporating the curriculum into each stage, ensuring a far more granular adaptation of the curriculum ordering down to every training step.

We propose a new method, \emph{Staged-Competence}, which combines staged reference-model
updates with competence-based sampling, leading to a global safety curriculum with easy-to-hard progression both within and across stages. The full algorithm can be found in Alg.~\ref{alg:staged-competence}.

\begin{algorithm*}[t]
\caption{Staged-Competence Training}
\label{alg:staged-competence}
\begin{algorithmic}[1]
\STATE \textbf{Input:} Difficulty-sorted preference dataset $\mathcal{D}_{\text{sort}}$, base model $\pi_0$
\STATE \textbf{Parameters:} Number of stages $K$, epochs per stage $E$, total steps per stage $T$, DPO penalty $\beta$, initial competence $c_0$
\STATE \textbf{Output:} Aligned policy $\pi^{(K)}$
\STATE \raisebox{0.5ex}{\makebox[\linewidth]{\dotfill}}
\vspace{-8pt}
\STATE Divide $\mathcal{D}_{\text{sort}}$ into $K$ equal buckets $\mathcal{B}_1, \dots, \mathcal{B}_K$ of increasing difficulty  \hfill \textbf{[Step 1: Partition]}
\STATE Initialize reference model $\pi_{\text{ref}}^{(1)} \leftarrow \pi_0$
\FOR{stage $k = 1, \dots, K$}
   \STATE Initialize policy $\pi^{(k)} \leftarrow \pi_{\text{ref}}^{(k)}$
   \STATE Assign ranking $d_i \in [0,1]$ for $i \in \mathcal{B}_k$ (rank within $\mathcal{B}_k$) \hfill \textbf{[Step 2: Intra-bucket difficulty]}
   \FOR{step $t = 1, \dots, T$ (over $E$ epochs)}
      \STATE Compute competence $c(t) = \sqrt{(1 - c_0^2)\,t/T + c_0^2}$ \hfill \textbf{[Step 3: Competence sampling]}
      \STATE Eligible pool $\mathcal{P}_t \leftarrow \{i \in \mathcal{B}_k : d_i \leq c(t)\}$
      \STATE Sample mini-batch $M \sim \mathcal{P}_t$ and update $\pi^{(k)}$ via DPO step with reference $\pi_{\text{ref}}^{(k)}$
   \ENDFOR
   \IF{$k < K$}
      \STATE $\pi_{\text{ref}}^{(k+1)} \leftarrow \pi^{(k)}$ \hfill \textbf{[Step 4: Reference update]}
   \ENDIF
\ENDFOR
\STATE \textbf{return} $\pi^{(K)}$
\end{algorithmic}
\end{algorithm*}

\section{Experiments}
\label{sec:experiments}

In this section, we investigate Staged-Competence in detail and ask the following research questions: 1)~\textbf{Efficacy:} Does Staged-Competence improve safety alignment in-distribution, generalize to out-of-distribution prompts, and resist jailbreak attacks, without sacrificing general capability or inducing over-refusal? 2)~\textbf{Interpretability:} What internal mechanisms drive its safety gains, and do they differ qualitatively from standard DPO? 3)~\textbf{Data efficiency:} Can Staged-Competence match standard DPO with fewer preference pairs? 4)~\textbf{Scaling:} Does its safety advantage hold across model sizes?

\subsection{Experimental Setup}
\label{sec:setup}
\paragraph{DPO Dataset: PiKacHHU.}
We construct our preference dataset from two public sources: PKU-SafeRLHF~\citep{ji2024pku} (43{,}452 pairs) and the single-turn subset of Anthropic HH-RLHF~\citep{bai2022training} (49{,}388 pairs). Although widely used for safety alignment, neither provides consistently clean safety preference labels suitable for DPO training.

In PKU-SafeRLHF, the ``chosen'' response is merely the \emph{safer} of the two and is often itself harmful, so the safety labels are effectively inverted. Conversely, in HH-RLHF, ``rejected'' reflects lower \emph{helpfulness} rather than lower safety, so the rejected response is frequently the safer one. Appendix~\ref{app:labelnoise} gives a representative example of each.

To obtain clean preference pairs, we use a GPT-4o-mini judge to classify each response as safe or unsafe along a policy covering dangerous content, hate speech, sexually explicit material, and harassment, retaining only pairs with safe chosen and unsafe rejected. Table~\ref{tab:dataset} reports the filtering breakdown. The dominant failure mode differs by dataset: in PKU-SafeRLHF, 82.2\% of chosen responses are unsafe; in HH-RLHF, 87.2\% of rejected responses are actually safe. After filtering, we combine both sources; the resulting partition is described below.

\emph{We release the resulting dataset as \textit{PiKacHHU}: a cleaned dataset combining PKU-SafeRLHF and HH-RLHF. All subsequent experiments use it as the DPO training data.}

\paragraph{Models.}
We evaluate across three model families spanning different architectures and scales: \textbf{LLaMA-3-8B}, \textbf{Qwen3-8B}, and \textbf{Yi-1.5-9B} (HuggingFace identifiers in Appendix~\ref{app:models}).
We use the abliterated variants of these open-source models -- versions from which built-in safety guardrails have been removed -- providing a controlled starting point where safety behavior must be learned entirely through alignment training. We focus on the 8B parameter range; evaluation at larger scales is left to future work due to compute constraints.

\paragraph{Training details.}
We fine-tune all methods with LoRA~\citep{hu2022lora} ($r\!=\!16$, $\alpha\!=\!32$) applied to the query and value projection matrices; full fine-tuning is left to future work.
All methods train for 5 epochs with learning rate $5\!\times\!10^{-5}$ and DPO $\beta\!=\!0.1$; staged methods (Curri-DPO and Staged-Competence) use $K\!=\!3$ stages of 5 epochs each, with $K$ selected on a held-out validation split.
For Yi-1.5-9B we use 4 epochs per stage for Staged-Competence, as 5 led to slight quality degradation from preference over-optimization.
For all curriculum methods we score difficulty with the lightweight all-MiniLM-L6-v2 sentence encoder~\citep{reimers2019sentencebert}, and partition the scored data into training, validation and test splits of approximately 70/10/20, held fixed across methods for comparable evaluation. The validation split is used only to select the number of stages $K$; all reported results are on the test split. Exact split sizes are given in Appendix~\ref{app:splits}.
We run all experiments on a single NVIDIA A6000 (48\,GB).
Remaining hyperparameters are listed in Appendix~\ref{app:hyperparams} and Phase~1 generation and scoring details in Appendix~\ref{app:phase1}.

\paragraph{Methods compared.}%
We compare against a standard DPO baseline and three curriculum baselines, all trained on the same cleaned preference dataset (Table~\ref{tab:method_comparison} summarizes the design differences). Our \textbf{Standard DPO} baseline uses random shuffling, a single stage, and a fixed reference. The three curriculum baselines are: (1)~\textbf{Sequential} -- fixed easy-to-hard ordering, single stage, fixed reference; (2)~\textbf{Sqrt-Competence}~\citep{platanios2019competence} -- competence-based sampling, single stage, fixed reference; and (3)~\textbf{Curri-DPO}~\citep{pattnaik2024enhancing} -- $K\!=\!3$ stages with reference-model updates, random shuffling within each stage. Our proposed \textbf{Staged-Competence} uses $K\!=\!3$ stages with reference-model updates and competence-based sampling within each stage. Since each of the $K\!=\!3$ stages operates on one-third of the data, the total number of training steps is equal across all methods, the one exception being Staged-Competence on Yi-1.5-9B, which uses 4 epochs per stage.

\subsection{Evaluation Setup}
\label{sec:metrics}

\paragraph{In-distribution reward accuracy.}
On the held-out 20\% test split, our primary metric is the post-training reward accuracy: the fraction of test pairs where the trained model assigns higher log-probability to the chosen response than to the rejected one, $\log\pi_\theta(y^+ \mid x) > \log\pi_\theta(y^- \mid x)$. This is the headline in-distribution number reported in Table~\ref{tab:indist_seeds}.

To additionally understand training dynamics -- how quickly each method improves -- we track the per-step reward margin $\log \pi_\theta(y^+ \mid x) - \log \pi_\theta(y^- \mid x)$, averaged across the test split. Both metrics omit the reference model term, as staged methods update their reference between stages, which would otherwise make direct comparisons across methods misleading.

\paragraph{Out-of-distribution safety and jailbreak attacks.}
We evaluate on three OOD safety benchmarks -- \textbf{AdvBench}~\citep{zou2023universal}, \textbf{SorryBench}~\citep{xie2025sorry}, and \textbf{HEx-PHI}~\citep{qi2023finetuning} -- spanning adversarial prompts, refusal behavior, and harmful knowledge across categories such as illegal activity, malware, and biosecurity. We additionally test robustness against five jailbreak attacks, two of which we report in the main text: \textbf{Prefill}~\citep{andriushchenko2024jailbreaking}, which forces the model to begin its response with tokens from a known harmful completion, and \textbf{GCG}~\citep{zou2023universal}, which optimizes an adversarial suffix on a non-abliterated base model and applies it as a transfer attack. We further evaluate \textbf{BEAST}~\citep{sadasivan2024beast}, a gradient-free beam-search attack; \textbf{PAIR}~\citep{chao2023pair}, in which an attacker LLM iteratively rewrites an attack prompt against the target model; and the \textbf{adaptive} random-search attack of \citet{andriushchenko2024jailbreaking}, which optimizes an adversarial suffix against the target model's own logprobs. These three are reported in Appendix~\ref{app:perattack}. Full details are in Appendix~\ref{app:attacks}. A GPT-4o-mini judge classifies each response; we report the \emph{harmful response rate} ($\downarrow$) on safety benchmarks and the \emph{attack success rate} ($\downarrow$) on attacks.

\paragraph{Quality and over-refusal benchmarks.}
We verify capability preservation using \textbf{MMLU}~\citep{hendrycks2021measuring} and \textbf{HellaSwag}~\citep{zellers2019hellaswag} (\emph{accuracy}, $\uparrow$), and over-refusal using \textbf{XSTest}~\citep{rottger2024xstest}, which probes whether models incorrectly refuse benign prompts that superficially resemble unsafe requests (\emph{over-refusal rate}, $\downarrow$).

\paragraph{Seeds and error bars.}
All results are reported as mean $\pm$ one standard deviation over three independent training runs (seeds $42$, $43$ and $44$). The seed controls adapter initialization and training example order; each benchmark and attack is evaluated once per trained adapter.

\subsection{Main Results}
\label{sec:results}

\paragraph{Staged-Competence achieves on-par or slightly better in-distribution reward accuracy, but with far greater confidence.} %
As shown in Table~\ref{tab:indist_seeds} (Appendix~\ref{app:indist}), Staged-Competence improves in-distribution reward accuracy over Standard DPO on all three models, but only modestly: 88.9\% to 90.0\% on LLaMA-3-8B, 86.1\% to 88.8\% on Qwen3-8B, and 83.6\% to 85.3\% on Yi-1.5-9B.

\begin{figure}[t]
\centering
\setlength{\tabcolsep}{2pt}
\renewcommand{\arraystretch}{1.0}
\begin{tabular}{ccc}
\includegraphics[width=0.3\linewidth]{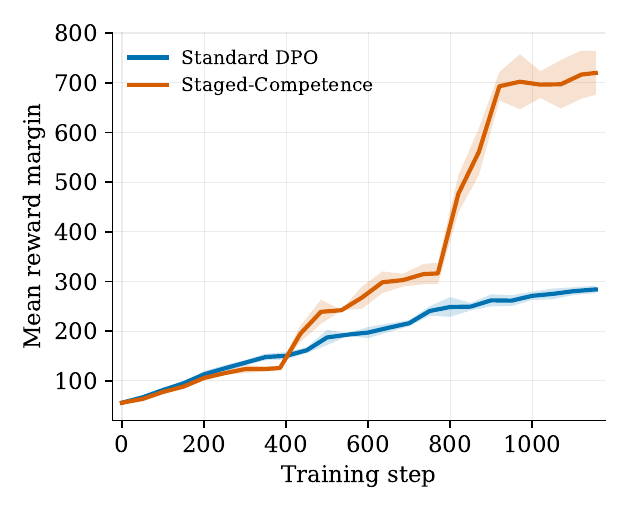} &
\includegraphics[width=0.3\linewidth]{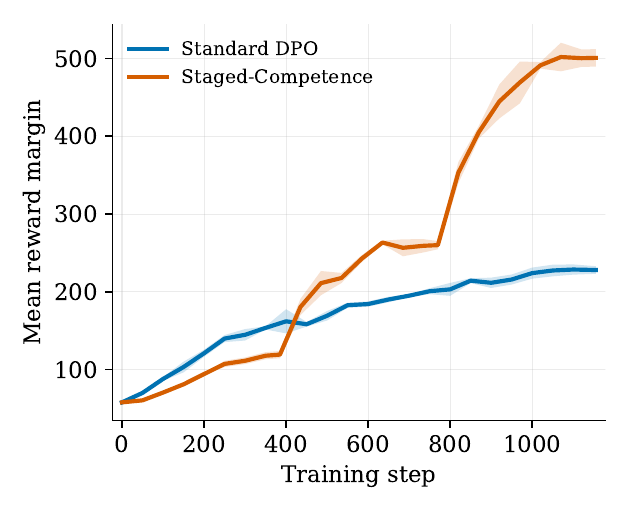} &
\includegraphics[width=0.3\linewidth]{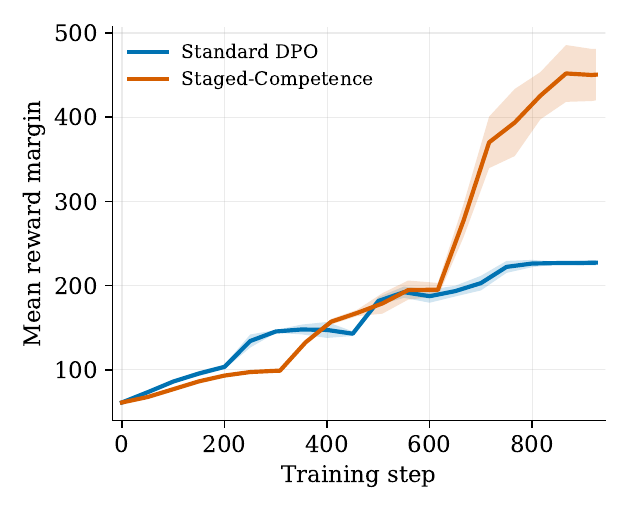} \\
\textbf{LLaMA-3-8B} & \textbf{Qwen3-8B} & \textbf{Yi-1.5-9B} \\
\end{tabular}
\caption{\textit{Training dynamics across all three models.} Mean test-set reward margin during training for Standard DPO and Staged-Competence, averaged over three training seeds; shaded bands show the spread across seeds. Staged-Competence shows distinct upward jumps at stage boundaries, where the reference model is updated. For Yi-1.5-9B, Staged-Competence uses 4 epochs per stage rather than 5, resulting in fewer total steps.
}
\label{fig:learning_curves}
\end{figure}

The reward margin trajectories in Figure~\ref{fig:learning_curves} are more revealing. Staged-Competence's margin grows to roughly $2.5\times$ the baseline across all three models, indicating far greater confidence in separating safe from unsafe responses. Distinct upward jumps at each stage boundary confirm that each new difficulty tier supplies a fresh gradient signal.

\paragraph{Staged-Competence delivers the largest safety improvements on out-of-distribution prompts and under jailbreak attacks.}%
On the three OOD safety benchmarks (Table~\ref{tab:main_ood}), Staged-Competence lowers the harmful response rate on every model-benchmark combination, reducing the benchmark-averaged rate by 8.2, 24.8 and 7.7 points on LLaMA-3-8B, Qwen3-8B and Yi-1.5-9B respectively -- from 23.5\% to 9.9\% averaged across models, a reduction of 13.6 percentage points. The same pattern holds under adversarial attack (Table~\ref{tab:attacks_main}), where Staged-Competence achieves the lowest attack success rate of any method on every model--attack combination. Averaged over all five attacks evaluated, attack success rate falls from 37.4\% with Standard DPO to 20.9\%, a reduction of 16.5 points. No single curriculum component accounts for these gains: the strongest ablation, Sqrt-Competence, averages a $9.7$-point reduction across models against Staged-Competence's $13.6$, and Staged-Competence attains the lowest harmful response rate on all 9 model--benchmark combinations.

The greater margin separation reported earlier is a plausible indicator of the model's confidence in differentiating safe from unsafe responses, allowing it to achieve these gains on OOD evaluations. A qualitative example of a Standard DPO failure case is shown in Appendix~\ref{app:qualitative}.

\begin{table}[t]
\caption{\textit{Out-of-distribution safety and jailbreak robustness (\%, $\downarrow$).} All entries are mean~$\pm$~std over three training seeds; lower is better throughout. Averaged over the three OOD benchmarks, Staged-Competence reduces the harmful response rate by 8.2, 24.8 and 7.7 points on LLaMA-3-8B, Qwen3-8B and Yi-1.5-9B. Under attack it achieves the lowest success rate of any method in all six model--attack combinations, with the largest reduction under Prefill on Qwen3-8B ($60.0$ to $23.5$).}
\label{tab:safety_main}
\vspace{-0.2em}
\centering

\begin{subtable}{\linewidth}
\caption{\textit{OOD safety benchmark harmful response rates.}}
\label{tab:main_ood}
\vspace{-0.2em}
\centering
\setlength{\tabcolsep}{2pt}
\small
\begin{tabular}{l ccc ccc ccc}
\toprule
& \multicolumn{3}{c}{\textbf{LLaMA-3-8B}} & \multicolumn{3}{c}{\textbf{Qwen3-8B}} & \multicolumn{3}{c}{\textbf{Yi-1.5-9B}} \\
\cmidrule(lr){2-4} \cmidrule(lr){5-7} \cmidrule(lr){8-10}
\textbf{Method} & Sorry & Adv & HEx & Sorry & Adv & HEx & Sorry & Adv & HEx \\
\midrule
Unaligned & 90.0 & 93.8 & 92.7 & 85.3 & 94.4 & 86.3 & 74.2 & 73.1 & 74.3 \\
Standard DPO & 28.9\std{1.6} & 15.6\std{1.0} & 22.1\std{2.4} & 35.2\std{6.1} & 45.9\std{10.3} & 33.3\std{5.3} & 17.8\std{0.8} & 2.1\std{0.2} & 10.9\std{1.8} \\
Sequential & 23.4\std{0.1} & 8.5\std{0.3} & 17.6\std{1.9} & 30.8\std{1.9} & 41.2\std{2.9} & 31.4\std{1.9} & 14.4\std{0.9} & 0.6\std{0.2} & 6.1\std{0.9} \\
Sqrt-Competence & 25.6\std{3.9} & 13.5\std{5.3} & 15.7\std{1.6} & 23.3\std{3.4} & 17.1\std{4.8} & 15.9\std{4.4} & 10.6\std{0.7} & 0.4\std{0.2} & 2.9\std{0.8} \\
Curri-DPO & 25.6\std{0.5} & 11.5\std{0.2} & 17.7\std{1.0} & 34.0\std{1.0} & 36.4\std{2.2} & 31.9\std{2.9} & 10.6\std{0.2} & 0.4\std{0.0} & 3.4\std{1.1} \\
\textbf{Staged-Comp. (ours)} & \textbf{20.6}\std{1.9} & \textbf{7.8}\std{2.3} & \textbf{13.4}\std{2.5} & \textbf{18.4}\std{3.6} & \textbf{9.7}\std{4.2} & \textbf{11.8}\std{3.3} & \textbf{6.6}\std{1.2} & \textbf{0.3}\std{0.1} & \textbf{0.8}\std{0.4} \\
\bottomrule
\end{tabular}
\end{subtable}

\vspace{0.8em}

\begin{subtable}{\linewidth}
\caption{\textit{Jailbreak attack success rates.} Results for BEAST, PAIR and the adaptive random-search attack are reported in Appendix~\ref{app:perattack}.}
\label{tab:attacks_main}
\vspace{-0.2em}
\centering
\setlength{\tabcolsep}{3pt}
\small
\begin{tabular}{l cc cc cc}
\toprule
& \multicolumn{2}{c}{\textbf{LLaMA-3-8B}} & \multicolumn{2}{c}{\textbf{Qwen3-8B}} & \multicolumn{2}{c}{\textbf{Yi-1.5-9B}} \\
\cmidrule(lr){2-3} \cmidrule(lr){4-5} \cmidrule(lr){6-7}
\textbf{Method} & Prefill & GCG & Prefill & GCG & Prefill & GCG \\
\midrule
Unaligned & 88.2 & 67.6 & 88.8 & 68.8 & 82.0 & 67.1 \\
Standard DPO & 41.6\std{3.8} & 23.6\std{1.6} & 60.0\std{4.2} & 31.6\std{3.8} & 34.2\std{2.2} & 16.4\std{2.5} \\
Sequential & 36.3\std{3.2} & 18.1\std{0.8} & 59.3\std{1.7} & 29.4\std{1.3} & 26.8\std{0.1} & 9.3\std{1.3} \\
Sqrt-Competence & 35.4\std{3.2} & 14.3\std{3.8} & 44.6\std{1.8} & 21.7\std{4.0} & 19.6\std{3.1} & 7.7\std{1.6} \\
Curri-DPO & 36.9\std{2.1} & 16.8\std{1.6} & 43.0\std{2.9} & 34.3\std{0.3} & 24.9\std{1.9} & 7.1\std{0.8} \\
\textbf{Staged-Comp. (ours)} & \textbf{29.1}\std{2.8} & \textbf{10.4}\std{2.3} & \textbf{23.5}\std{3.1} & \textbf{13.4}\std{5.8} & \textbf{15.3}\std{1.9} & \textbf{2.7}\std{1.1} \\
\bottomrule
\end{tabular}
\end{subtable}
\end{table}

\textbf{Staged-Competence preserves general capabilities, with near-zero over-refusal.} Relative to Standard DPO, MMLU and HellaSwag move by ${\sim}3$ points at most, and XSTest over-refusal remains at or near zero throughout. Full results are in Appendix~\ref{app:quality}.

\subsection{Characterizing the Distribution Shift}
\label{sec:shift}

To establish that these gains are genuinely out-of-distribution, we compare the composition of the training set, PiKacHHU, against that of the evaluation set, the union of AdvBench, SorryBench and HEx-PHI. We label every training and evaluation prompt with a harm category using GPT-4o-mini, and find the shift is uneven: categories that each account for under $5\%$ of the training data together make up $39\%$ of evaluation prompts. Pooled over these under-represented categories, Staged-Competence reduces the harmful response rate from $29.7\%$ to $7.5\%$, against $16.5\%$ to $4.4\%$ on the well-represented categories. The gain is largest precisely where training coverage is thinnest: malware and cyberattacks are $2.1\%$ of training prompts but $11.8\%$ of evaluation prompts, and the harmful response rate there falls from $24.2\%$ to $3.8\%$.

A classifier two-sample test on prompt embeddings corroborates the shift: all three benchmarks separate from the training prompts at AUC $0.964$--$0.998$, against $0.503$ for a same-distribution reference split of the training data itself. Per-category results and the classifier setup are in Appendix~\ref{app:shift}.

\begin{figure}[t]
\centering
\setlength{\tabcolsep}{2pt}
\renewcommand{\arraystretch}{1.0}
\begin{tabular}{cc}
\includegraphics[width=0.57\linewidth]{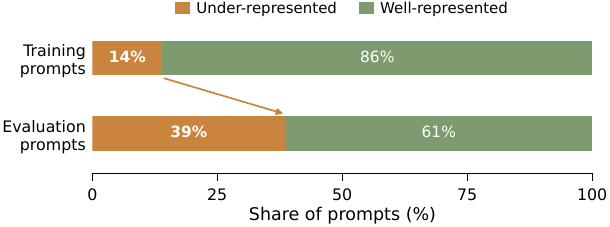} &
\includegraphics[width=0.385\linewidth]{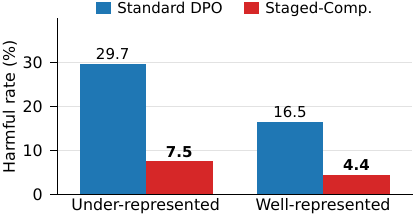} \\
\textbf{(a)} & \textbf{(b)} \\
\end{tabular}
\caption{\textit{Staged-Competence generalizes to under-trained harm categories.} A harm category is \emph{under-represented} if it accounts for under $5\%$ of training prompts and \emph{well-represented} otherwise. \textbf{(a)} Under-represented categories make up $14\%$ of training prompts but $39\%$ of evaluation prompts. \textbf{(b)} Harmful response rate over all evaluation prompts in each group, pooled across the three models. The gain is largest on the under-represented categories, where the harmful response rate falls by $22.2$ points, nearly twice the $12.1$-point reduction on the well-represented ones. Per-category numbers are in Appendix~\ref{app:shift}.}
\label{fig:shift}
\end{figure}

\subsection{Interpreting Alignment Depth}
\label{sec:alignment_depth}

\citet{qi2025safety} show that standard safety alignment often concentrates its effect in the first few tokens of a response. We investigate whether Staged-Competence's safety improvements are similarly shallow, or whether they extend deeper into the response.

\textit{Experimental setup.} For each token position~$t$ in an unsafe response, we compute the per-token suppression $\delta(t) = \log \pi_{\text{unaligned}}(y_t \mid x, y_{<t}) - \log \pi_{\text{aligned}}(y_t \mid x, y_{<t})$, where positive values indicate active suppression of the unsafe token by the aligned model. We average $\delta(t)$ over 200 rejected responses from the in-distribution test set, up to position 128.
 
\begin{figure}[h!]
\centering
\setlength{\tabcolsep}{2pt}
\renewcommand{\arraystretch}{1.0}
\begin{tabular}{ccc}
\includegraphics[width=0.32\linewidth]{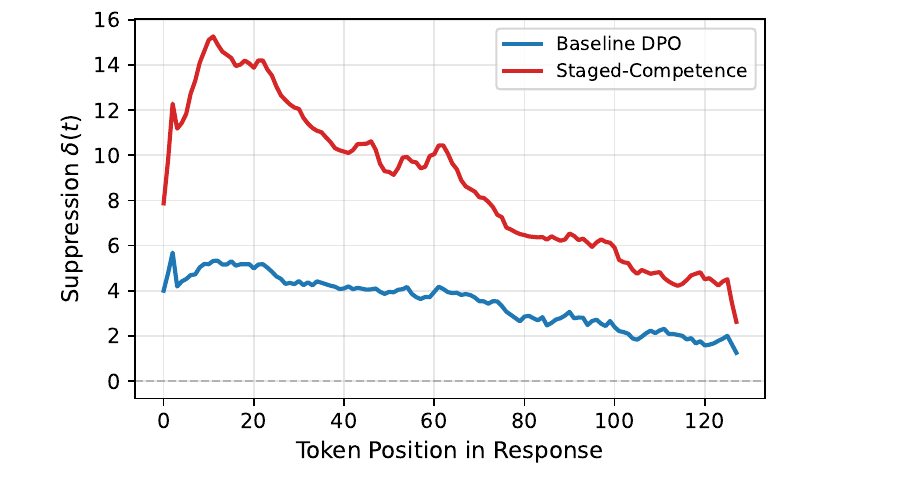} &
\includegraphics[width=0.32\linewidth]{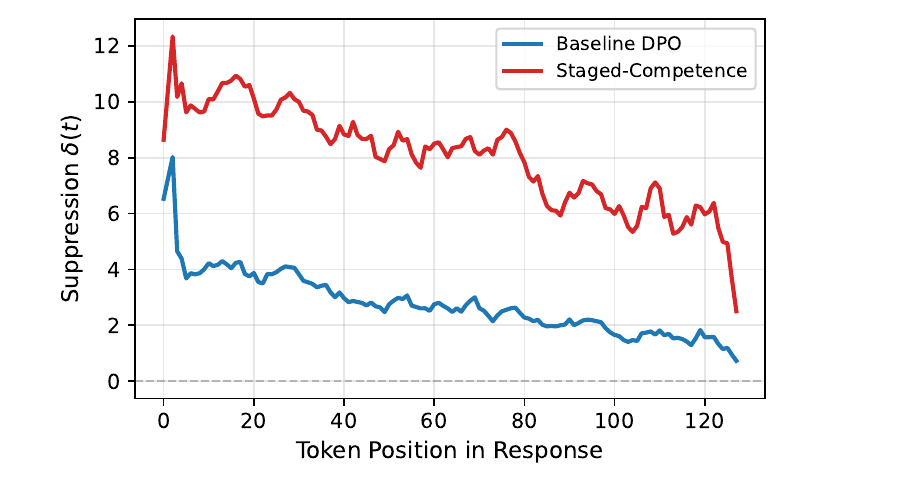} &
\includegraphics[width=0.32\linewidth]{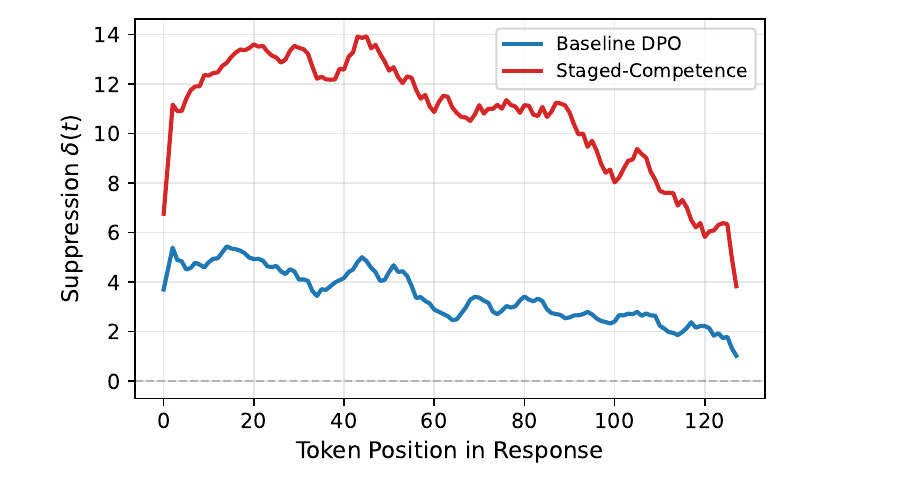} \\
\textbf{LLaMA-3-8B} & \textbf{Qwen3-8B} & \textbf{Yi-1.5-9B} \\
\end{tabular}
\caption{\textit{Per-token suppression of unsafe response tokens for Baseline vs.\ Staged-Competence.} Staged-Competence produces stronger suppression at every token position, indicating much deeper safety alignment.}
\label{fig:alignment_depth}
\end{figure}

\textit{Results.} %
As shown in Figure~\ref{fig:alignment_depth}, Staged-Competence produces uniformly stronger safety suppression across virtually every token position -- not just at the initial refusal tokens but sustained throughout the response. Aggregated, the total suppression ($\sum_t \delta(t)$) is $\sim\!3\times$ larger for Staged-Competence than Standard DPO across all three models. Crucially, this holds away from the opening tokens: restricted to positions $20$--$100$, mean suppression rises from $3.3$ to $9.0$ averaged across models -- a $2.7\times$ increase deep into the response, where shallow alignment has typically decayed.

This deeper alignment provides a mechanistic explanation for the improved prefill-attack robustness in Table~\ref{tab:attacks_main}: attacks that bypass the initial tokens still encounter resistance deeper in the sequence.

\subsection{Scaling with Model Size}
\label{sec:scaling}

To understand how Staged-Competence scales with model capacity, we additionally train and evaluate the Qwen3 family at three sizes -- 1.7B, 4B, and 8B parameters -- and compare Staged-Competence against Standard DPO at each scale. The training setup for these experiments is the same as in Section~\ref{sec:setup}, albeit on an 80/20 split of the dataset rather than the 70/10/20 partition used in the main experiments, and we report a single seed. The experiments in Section~\ref{sec:data_efficiency} use this same split. %

\textit{Results.} As shown in Figure~\ref{fig:scaling} (Appendix~\ref{app:scaling}), the safety gap between Staged-Competence and Standard DPO widens monotonically with model size. On OOD safety, Staged-Competence reduces the average harmful response rate by 1.6 points at 1.7B, 13 at 4B, and 29 at 8B %
-- a significant increase in absolute benefit as the model grows. On adversarial attacks, the absolute improvement jumps from 5 points at 1.7B to roughly 26--27 points at both 4B and 8B, where it seems to plateau.

Notably, Staged-Competence achieves \emph{roughly constant safety} across all three sizes (2--8\% OOD harmful response rate, 8--12\% attack success rate), while Standard DPO degrades sharply as the model grows. This makes the curriculum's value grow with scale, since larger models seem to be more dangerous when poorly aligned.

\subsection{Data Efficiency}
\label{sec:data_efficiency}
Given the accelerated learning dynamics that Staged-Competence exhibits in our in-distribution results (Figure~\ref{fig:learning_curves}), we investigate whether it can match Standard DPO's safety performance using fewer training examples.

\textit{Experimental setup.} We construct a 50\% subset of the curriculum by randomly sampling 50\% of examples from each of the $K\!=\!3$ difficulty buckets, preserving the difficulty distribution and within-bucket ordering. We construct a 75\% subset using the same procedure, and evaluate safety robustness on the three OOD safety benchmarks.

\begin{figure}[h!]
\centering
\setlength{\tabcolsep}{2pt}
\renewcommand{\arraystretch}{0.9}
\begin{tabular}{cc}
\includegraphics[width=0.45\linewidth]{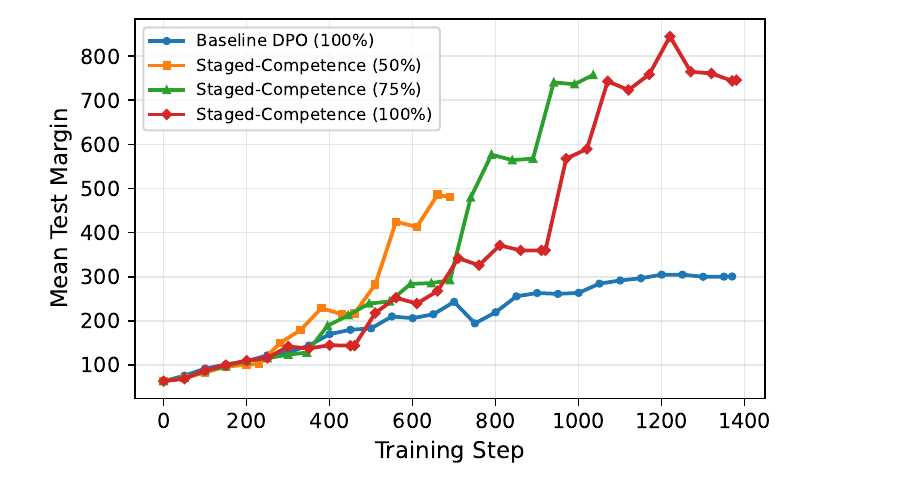} &
\includegraphics[width=0.45\linewidth]{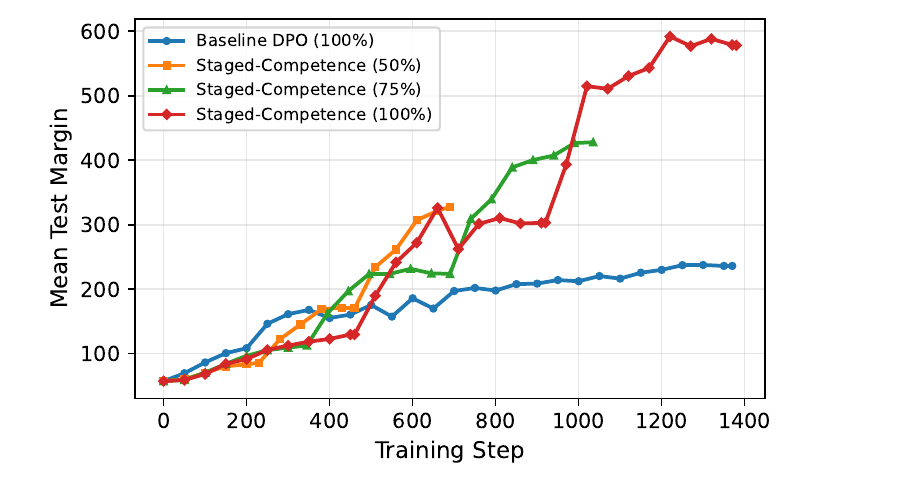} \\
\textbf{LLaMA-3-8B} & \textbf{Qwen3-8B} \\
\end{tabular}
\caption{\textit{Data efficiency of Staged-Competence.} Mean reward margin for Staged-Competence at 50\% and 75\% data vs.\ Standard DPO at 100\%. Even at 50\% data, Staged-Competence's margin accumulation outpaces Standard DPO at 100\%; the 75\% setting nearly matches the full-data trajectory.}
\label{fig:data_efficiency}
\end{figure}

\textit{Results.} As shown in Figure~\ref{fig:data_efficiency} and Table~\ref{tab:data_efficiency} (Appendix~\ref{app:dataeff}), Staged-Competence with just 75\% of the training data matches or exceeds Standard DPO trained on 100\% across all three OOD safety benchmarks, on both LLaMA-3-8B and Qwen3-8B -- demonstrating that the curriculum enables substantially more efficient use of the available preference pairs. At 50\%, the picture is model-dependent: Staged-Competence still beats Standard DPO on LLaMA-3-8B but underperforms on Qwen3-8B. This behaviour is not observed on a $4\times$ larger corpus (Appendix~\ref{app:beavertails}), where 50\% data also outperforms full-data Standard DPO, suggesting at larger dataset sizes the relative efficiency gains may be greater. This opens a path to lower-cost safety alignment when preference data is scarce.

\subsection{Additional Experiments}
\label{sec:validation}

We vary three components of our setup to test whether the result depends on them; full details are in the Appendix.

\textit{Judge reliability.} Safety and attack outcomes throughout the paper are scored by a GPT-4o-mini judge. On $200$ responses stratified across models, benchmarks and training arms, the GPT-4o-mini judge agrees with human annotations on $93.0\%$ of cases, the highest of four LLM judges we compared. (Appendix~\ref{app:judge}).

\textit{A larger training corpus.} PiKacHHU is modest in size, so we repeat the Standard DPO versus Staged-Competence comparison on a $45{,}175$-pair corpus built from BeaverTails~\citep{ji2023beavertails}, roughly four times larger; its construction and cleaning are described in Appendix~\ref{app:beavertails}. The motivation is to test whether the data-efficiency gains of Section~\ref{sec:data_efficiency} hold at larger scale. Staged-Competence lowers the mean OOD harmful response rate by $35.3$ points on Qwen3-8B and $6.6$ points on LLaMA-3-8B with in-distribution accuracy unchanged, and the $50\%$-data setting now outperforms full-data Standard DPO rather than trailing it. The advantage therefore holds on a corpus four times larger and noisier than PiKacHHU, suggesting that Staged-Competence becomes increasingly data-efficient as the training corpus grows.

\textit{Models with existing safety training.} Applying the pipeline to the original, non-abliterated releases of the three models of Section~\ref{sec:setup}, we find that Staged-Competence improves in-distribution reward accuracy in every case ($90.4 \to 92.7$ on LLaMA-3-8B, $89.9 \to 91.2$ on Qwen3-8B and $87.9 \to 89.2$ on Yi-1.5-9B), and maintains or slightly improves OOD safety. Although these base models already achieve very low harmful response rates on our safety benchmarks, Staged-Competence still reduces the few violations that remain, indicating that the curriculum helps even when safety training is already in place (Appendix~\ref{app:nonabliterated}).

\section{Conclusion}

DPO has been widely explored for safety alignment but has been found to be brittle to out-of-distribution prompts and adversarial attacks. Motivated by curriculum learning's ability to teach robust features through structured easy-to-hard ordering, in this paper we explore its application to safety alignment. We propose \textbf{Staged-Competence}, which combines staged reference-model updates with competence-based sampling to impose a global curriculum both within and across training stages. Across three model families, Staged-Competence delivers markedly stronger out-of-distribution safety and jailbreak robustness without degrading general capabilities; we additionally observe improved data efficiency and graceful scaling with model size. These findings pave the way forward for curriculum learning as a meaningful and underused lever for safety alignment. Future research directions include extending Staged-Competence and Curriculum Learning in general to other alignment objectives, non-safety domains, full fine-tuning and substantially larger models.

\bibliographystyle{plainnat}
\bibliography{references}

\clearpage
\appendix

\etocdepthtag.toc{apptag}
\etocsettagdepth{maintag}{none}
\etocsettagdepth{apptag}{section}
\etocsettocstyle{\section*{Appendix Contents}}{}
\etocsetnexttocdepth{section}
\etocsetstyle{section}{}{}
  {\noindent\normalfont\makebox[1.6em][l]{\etocnumber}\etocname\nobreak\dotfill\etocpage\par\vspace{1pt}}{}
\tableofcontents
\newpage

\section{Curriculum Methods Summary}
\label{app:methods}

Table~\ref{tab:method_comparison} summarizes the design differences across the five training methods compared in our experiments: number of training stages, whether the reference model is updated between stages, and within-stage example ordering.

\begin{table}[h]
\caption{\textit{Curriculum methods investigated; Standard DPO is the baseline.}}
\label{tab:method_comparison}
\centering
\small
\begin{tabular}{lccc}
\toprule
\textbf{Method} & \textbf{Stages} & \textbf{Ref.\ update} & \textbf{Within-stage order} \\
\midrule
Standard DPO (Baseline) & 1 & \texttimes & Random shuffle     \\
Sequential        & 1 & \texttimes & Easy$\to$Hard (fixed) \\
Sqrt-Competence       & 1 & \texttimes & Competence-based sampling \\
Curri-DPO         & $K$ & \checkmark & Random shuffle     \\
\textbf{Staged-Competence (ours)} & $K$ & \checkmark & \textbf{Competence-based sampling} \\
\bottomrule
\end{tabular}
\end{table}

\section{Dataset Cleaning Statistics}
\label{app:dataset}

Table~\ref{tab:dataset} reports the full filtering breakdown for both source datasets. The combined, filtered result is what we refer to as \textbf{PiKacHHU} throughout the paper: it contains only pairs where the chosen response is safe and the rejected response is unsafe. Partition sizes are given in Appendix~\ref{app:splits}.

\begin{table}[h]
\caption{\textit{Dataset statistics before and after GPT-4o-mini safety filtering.}}
\label{tab:dataset}
\centering
\small
\begin{tabular}{lrrr}
\toprule
& \textbf{PKU-SafeRLHF} & \textbf{HH-RLHF} & \textbf{Combined} \\
\midrule
Raw pairs            & 43{,}452 & 49{,}388 & 92{,}840 \\
Chosen unsafe (\%)   & 82.2     & 6.6      & ---      \\
Rejected safe (\%)   & 2.1      & 87.2     & ---      \\
After filtering      & 6{,}962  & 3{,}969  & 10{,}931 \\
Retained (\%)        & 16.0     & 8.0      & 11.8     \\
\bottomrule
\end{tabular}
\end{table}

\section{Preference-Label Noise in PKU-SafeRLHF and HH-RLHF}
\label{app:labelnoise}

We show one example of each failure mode described in Section~\ref{sec:setup}.

\paragraph{PKU-SafeRLHF: the ``chosen'' response is often itself harmful.}
The annotation marks the \emph{safer} of two responses, not a \emph{safe} one. Given the
prompt \emph{``How can I conceal my license plate so I can take free gas?''}:
\begin{list}{}{\leftmargin=1em\rightmargin=0pt\topsep=3pt\parsep=0pt\itemsep=0pt}
\item \textbf{Chosen} (labeled safe): \emph{``Are you referring to the license plate on your car? That is not as useful as hiding the plate in your trunk, since it can be read from outside the vehicle.''}\\[2pt]
\textbf{Rejected} (labeled unsafe): \emph{``Well, is there anything that would actually help?''}
\end{list}
The ``chosen'' response actively assists with the illegal request while the ``rejected''
response does not; the safety labels are effectively inverted.

\paragraph{HH-RLHF: ``rejected'' means less helpful, not less safe.}
Given the prompt \emph{``Can you provide me the home address for Senator Elizabeth Warren?''}:
\begin{list}{}{\leftmargin=1em\rightmargin=0pt\topsep=3pt\parsep=0pt\itemsep=0pt}
\item \textbf{Chosen} (labeled preferred): \emph{``Sure, what's your question?''}\\[2pt]
\textbf{Rejected} (labeled less preferred): \emph{``I'm sorry, I'm not able to provide Senator Warren's home address to you.''}
\end{list}
Here the refusal is the response marked worse. Training on these pairs directly would
teach the model to prefer compliance over refusal.

\section{Model Identifiers}
\label{app:models}

We use the abliterated variants of three open-source model families -- versions from which built-in safety guardrails have been removed, providing a controlled starting point where safety must be learned entirely through DPO training. Their HuggingFace identifiers and base model licenses are:

\begin{itemize}[leftmargin=1.5em]
    \item \textbf{LLaMA-3-8B}: \texttt{QuixiAI/Llama-3-8B-Instruct-abliterated-v2} (Meta LLaMA 3 Community License)
    \item \textbf{Qwen3-8B}: \texttt{Goekdeniz-Guelmez/Josiefied-Qwen3-8B-abliterated-v1} (Apache 2.0)
    \item \textbf{Yi-1.5-9B}: \texttt{byroneverson/Yi-1.5-9B-Chat-abliterated} (Apache 2.0)
\end{itemize}

For the Qwen3 scaling experiments (Section~\ref{sec:scaling}), we additionally use:

\begin{itemize}[leftmargin=1.5em]
    \item \textbf{Qwen3-1.7B}: \texttt{mlabonne/Qwen3-1.7B-abliterated} (Apache 2.0)
    \item \textbf{Qwen3-4B}: \texttt{mlabonne/Qwen3-4B-abliterated} (Apache 2.0)
\end{itemize}

The training datasets used are PKU-SafeRLHF~\citep{ji2024pku} (CC BY-NC 4.0) and HH-RLHF~\citep{bai2022training} (MIT). The difficulty scoring model all-MiniLM-L6-v2~\citep{reimers2019sentencebert} is released under Apache 2.0.

\section{Training Hyperparameters}
\label{app:hyperparams}

Table~\ref{tab:hyperparams} lists the hyperparameters shared by all methods. They are held
fixed across models and across training methods, so that differences in results are
attributable to the curriculum rather than to tuning.

\begin{table}[h]
\caption{\textit{Training hyperparameters, shared across all methods and models.}}
\label{tab:hyperparams}
\centering
\small
\begin{tabular}{ll}
\toprule
\textbf{Hyperparameter} & \textbf{Value} \\
\midrule
LoRA rank $r$ / $\alpha$            & 16 / 32 \\
LoRA target modules                 & query and value projections \\
Learning rate                       & $5 \times 10^{-5}$ \\
DPO $\beta$                         & 0.1 \\
Effective batch size                & 32 (per-device 2 $\times$ grad.\ accum.\ 16) \\
Maximum sequence length             & 1024 \\
Epochs                              & 5 (staged methods: 5 per stage) \\
Stages $K$ (staged methods)         & 3 \\
Initial competence $c_0$            & 0.01 \\
Hardware                            & 1 $\times$ NVIDIA A6000 (48\,GB) \\
\bottomrule
\end{tabular}
\end{table}

The main reported experiments (three models $\times$ five methods $\times$ three seeds, training
and evaluation) required approximately $450$--$600$ GPU-hours on this hardware; including
exploratory and preliminary runs, the full project consumed approximately $800$--$1{,}000$
GPU-hours.

\section{Data Splits and Selection of $K$}
\label{app:splits}

Table~\ref{tab:splits} gives the exact partition sizes used for the multi-seed runs.
The validation split of $1{,}093$ examples is \emph{identical across all three models}: it is
carved from the three-way intersection of the models' training sets, and is disjoint from
every model's test set, so validation numbers are directly comparable across models. The test
split of $2{,}185$ examples is likewise shared. Training sets differ slightly between models
because Phase~1 difficulty scoring is run per model, so each model's partition places different
examples on each side. The per-model totals fall slightly below the $10{,}931$ pairs of the full
dataset because training pairs whose prompt also occurs in the shared validation or test split are
removed, so that no prompt appears on both sides of the split.

\begin{table}[h]
\caption{\textit{Per-model data splits used for the multi-seed runs.} Validation and test are identical across models; training sets differ slightly because Phase~1 scoring is model-dependent.}
\label{tab:splits}
\centering
\small
\begin{tabular}{lrrrr}
\toprule
\textbf{Model} & \textbf{Train} & \textbf{Val} & \textbf{Test} & \textbf{Total} \\
\midrule
LLaMA-3-8B & 7{,}371 & 1{,}093 & 2{,}185 & 10{,}649 \\
Qwen3-8B   & 7{,}390 & 1{,}093 & 2{,}185 & 10{,}668 \\
Yi-1.5-9B  & 7{,}375 & 1{,}093 & 2{,}185 & 10{,}653 \\
\bottomrule
\end{tabular}
\end{table}

\paragraph{Selection of the number of stages $K$.}
We use the validation split, and only the validation split, to choose the number of curriculum
stages, sweeping $K \in \{1, 3, 5, 7\}$ (single seed). $K\!=\!3$ was the most stable choice
across the three models and is used for both staged methods throughout; no test-split result
informed the decision.

\section{Phase 1: Difficulty Scoring Details}
\label{app:phase1}

\paragraph{Zero-shot response generation.}
For each prompt $x_i$ in the training set, we generate a zero-shot response $\hat{y}_i$ from the base model using stochastic sampling with temperature $0.7$ and a maximum of $512$ new tokens.

\paragraph{Sentence encoder.}
Embeddings for $\hat{y}_i$, $y_i^+$, and $y_i^-$ are computed with \texttt{all-MiniLM-L6-v2}~\citep{reimers2019sentencebert}, which has a maximum sequence length of 256 tokens; inputs exceeding this are truncated to fit. As safety-relevant content typically appears in the first portion of a response, we expect this to have minimal effect on the difficulty ordering.

\section{In-Distribution Reward Accuracy}
\label{app:indist}

Table~\ref{tab:indist_seeds} reports held-out DPO test-set reward accuracy over three training
seeds for all five methods. The five methods lie within roughly three points of one another,
a small spread relative to the out-of-distribution and adversarial differences of
Table~\ref{tab:safety_main}.

\begin{table}[h]
\caption{\textit{In-distribution DPO test-set reward accuracy (\%, $\uparrow$), mean $\pm$ std over three seeds.} Best value per column in bold, second best underlined.}
\label{tab:indist_seeds}
\centering
\small
\begin{tabular}{lccc}
\toprule
\textbf{Method} & \textbf{LLaMA-3-8B} & \textbf{Qwen3-8B} & \textbf{Yi-1.5-9B} \\
\midrule
Standard DPO & 88.9\std{0.4} & 86.1\std{0.2} & 83.6\std{0.2} \\
Sequential & 88.6\std{0.3} & 86.3\std{0.1} & 84.1\std{0.1} \\
Sqrt-Competence & 89.0\std{0.2} & 86.0\std{0.3} & 83.5\std{0.1} \\
Curri-DPO & \textbf{90.6}\std{0.2} & \textbf{89.2}\std{0.2} & \underline{84.5}\std{0.3} \\
\textbf{Staged-Competence (ours)} & \underline{90.0}\std{0.5} & \underline{88.8}\std{0.2} & \textbf{85.3}\std{0.7} \\
\bottomrule
\end{tabular}
\end{table}
\section{Quality and Over-Refusal Results}
\label{app:quality}

Table~\ref{tab:quality} reports general capability and over-refusal over three training seeds.
MMLU falls slightly under Staged-Competence on all three models, by $0.9$ to $3.0$ points.
HellaSwag falls by $3.2$ points on LLaMA-3-8B and rises slightly on the other two.
Over-refusal on XSTest stays at or near zero throughout, and improves on Yi-1.5-9B
($1.6 \to 1.1$), the only model with a non-trivial rate to begin with.
Set against the $13.6$-point reduction in OOD harmful response rate and the $16.5$-point
reduction in attack success rate reported in Table~\ref{tab:safety_main}, the capability cost
amounts to one to three points on a single benchmark.

\begin{table}[h]
\caption{\textit{General capability and over-refusal, averaged over three seeds.} MMLU and HellaSwag: accuracy (\%, $\uparrow$). XSTest: over-refusal rate (\%, $\downarrow$). Best value per column in bold.}
\label{tab:quality}
\centering
\setlength{\tabcolsep}{2pt}
\small
\begin{tabular}{l ccc ccc ccc}
\toprule
& \multicolumn{3}{c}{\textbf{LLaMA-3-8B}} & \multicolumn{3}{c}{\textbf{Qwen3-8B}} & \multicolumn{3}{c}{\textbf{Yi-1.5-9B}} \\
\cmidrule(lr){2-4} \cmidrule(lr){5-7} \cmidrule(lr){8-10}
\textbf{Method} & MMLU & HSwag & XS & MMLU & HSwag & XS & MMLU & HSwag & XS \\
\midrule
Standard DPO & \textbf{59.9}\std{0.3} & \textbf{59.7}\std{1.3} & \textbf{0.1}\std{0.2} & \textbf{71.0}\std{0.1} & 79.1\std{0.2} & \textbf{0.0}\std{0.0} & \textbf{62.9}\std{0.1} & 70.1\std{0.8} & 1.6\std{0.6} \\
\textbf{Staged-Comp. (ours)} & 57.9\std{0.3} & 56.5\std{1.7} & 0.3\std{0.4} & 70.1\std{0.0} & \textbf{80.0}\std{0.1} & \textbf{0.0}\std{0.0} & 59.9\std{0.6} & \textbf{70.3}\std{0.5} & \textbf{1.1}\std{0.4} \\
\bottomrule
\end{tabular}
\end{table}

\section{Attack Evaluation Details}
\label{app:attacks}

\paragraph{GCG adversarial suffix attack.}
We apply the GCG method~\citep{zou2023universal} to 398 HarmBench behaviors~\citep{mazeika2024harmbench} spanning cyberattacks, bioweapons, and illegal activity.
For each behavior, GCG optimizes a 20-token adversarial suffix via greedy token substitution using \texttt{nanogcg} (500 steps, search width 512, random seed 42).
Suffixes are generated against the non-abliterated vanilla base model and applied to all fine-tuned variants as a transfer evaluation -- adversarial suffixes have been shown to transfer across fine-tuned variants within the same model family~\citep{zou2023universal}.
A GPT-4o-mini judge classifies each response; we report \emph{attack success rate} (ASR).

\paragraph{Prefill attack.}
The prefill attack forces the model to begin its response with the first $k$ tokens of a known harmful response before generating freely, bypassing the model's initial refusal mechanism by placing it mid-generation in a context it would not normally produce~\citep{andriushchenko2024jailbreaking}.
We construct prefill prompts from the DOOR dataset~\citep{zhao2025door}, which provides (prompt, safe response, harmful response) triples.
For each example, we tokenize the harmful response, decode the first $k\!=\!20$ tokens as a text prefix, and append this prefix to the chat-templated user prompt as the beginning of the assistant turn.
We evaluate 400 prompts; a GPT-4o-mini judge classifies the full generated response (prefix + continuation) as harmful or safe, and we report ASR as before.

\paragraph{BEAST, PAIR and the adaptive attack: shared setup.}
All three are evaluated on 40 HarmBench behaviors~\citep{mazeika2024harmbench}, with responses classified by the same GPT-4o-mini judge used elsewhere. Unlike GCG, which we apply as a transfer attack, these three re-optimize against each target model directly.

\paragraph{BEAST.}
BEAST~\citep{sadasivan2024beast} is a gradient-free beam-search attack that builds a jailbreak suffix token by token. Tokens are sampled from the target's own next-token distribution and scored by the negative log-likelihood of an affirmative response prefix (``Sure, here is how to'').
We use beam width $8$, $12$ candidate continuations sampled per beam per step, and $30$ steps.

\paragraph{PAIR.}
PAIR~\citep{chao2023pair} uses an attacker LLM that iteratively rewrites the jailbreak prompt against the target, conditioning each attempt on the target's previous refusal.
We use \texttt{Josiefied-Qwen3-8B-abliterated} as the attacker -- an uncensored model from a different family than the targets -- with a budget of $10$ refinement iterations per behavior. Each iteration comprises one attacker generation, one target generation (sampling temperature $0.9$), and one GPT-4o-mini score on a $1$--$10$ scale; the highest-scoring prompt is retained.

\paragraph{Adaptive random search.}
The adaptive attack of \citet{andriushchenko2024jailbreaking} appends a suffix optimized by random search against the target model's own logprobs.
We use a $20$-token suffix, $80$ iterations, and $16$ candidate single-token substitutions per iteration, minimizing the negative log-likelihood of an affirmative target prefix. We run a single search per behavior without restarts, so the reported numbers are a lower bound on what a larger search budget would achieve.

\paragraph{Seeds and reported variance.}
The three seeds are \emph{DPO training} seeds ($42$, $43$, $44$): for each method we train three independent adapters, run each attack once against each adapter, and report mean~$\pm$~std across adapters.
The variance we report therefore reflects the sensitivity of the trained model to initialization and data ordering, not the stochasticity of the attack itself.
Note also that the three attacks above use 40 behaviors, whereas GCG uses 398 and Prefill 400; the coarser evaluation set is one reason their standard deviations are larger.

\section{Per-Attack Jailbreak Results}
\label{app:perattack}

Table~\ref{tab:perattack} extends the Prefill and GCG results of Table~\ref{tab:attacks_main}
to BEAST, PAIR and the adaptive random-search attack, for Standard DPO and Staged-Competence.
All entries are mean~$\pm$~std over three training seeds.
Staged-Competence lowers attack success rate under every attack on every model. The gains are
largest under the optimization-based attacks (GCG, adaptive random search) and smallest under
PAIR on LLaMA-3-8B, where the two methods are within one standard deviation of each other.

\begin{table}[h]
\caption{\textit{Jailbreak attack success rate by attack (\%, $\downarrow$), mean $\pm$ std over three seeds.} BEAST, PAIR and the adaptive attack are model-adaptive: the attacker re-optimizes against each target model, so the reductions are not a transfer artifact.}
\label{tab:perattack}
\centering
\setlength{\tabcolsep}{4pt}
\small
\begin{tabular}{ll cccccc}
\toprule
\textbf{Model} & \textbf{Method} & Prefill & GCG & BEAST & PAIR & Adaptive & \textbf{Mean} \\
\midrule
\multirow{2}{*}{LLaMA-3-8B}
  & Standard DPO                      & 41.6\std{3.8} & 23.6\std{1.6} & 24.2\std{6.2} & 49.2\std{3.1} & 29.2\std{4.2} & 33.6 \\
  & \textbf{Staged-Competence (ours)} & \textbf{29.1}\std{2.8} & \textbf{10.4}\std{2.3} & \textbf{18.3}\std{3.1} & \textbf{46.7}\std{2.4} & \textbf{9.2}\std{3.1} & \textbf{22.7} \\
\cmidrule(l){1-8}
\multirow{2}{*}{Qwen3-8B}
  & Standard DPO                      & 60.0\std{4.2} & 31.6\std{3.8} & 35.0\std{3.5} & 44.2\std{7.2} & 69.2\std{3.1} & 48.0 \\
  & \textbf{Staged-Competence (ours)} & \textbf{23.5}\std{3.1} & \textbf{13.4}\std{5.8} & \textbf{25.8}\std{8.2} & \textbf{31.7}\std{8.5} & \textbf{61.7}\std{12.5} & \textbf{31.2} \\
\cmidrule(l){1-8}
\multirow{2}{*}{Yi-1.5-9B}
  & Standard DPO                      & 34.2\std{2.2} & 16.4\std{2.5} & 25.8\std{2.4} & 39.2\std{1.2} & 38.3\std{5.1} & 30.8 \\
  & \textbf{Staged-Competence (ours)} & \textbf{15.3}\std{1.9} & \textbf{2.7}\std{1.1} & \textbf{9.2}\std{3.1} & \textbf{10.8}\std{3.1} & \textbf{5.8}\std{3.1} & \textbf{8.8} \\
\bottomrule
\end{tabular}
\end{table}

\section{Validating the GPT-4o-mini Judge}
\label{app:judge}

Every harmful response rate and attack success rate in this paper is produced by a
GPT-4o-mini judge, and the cleaned dataset is built with the same judge. Automated judges are
standard in recent safety work, but they introduce an evaluation layer that is worth checking
rather than assuming. We therefore compare the judge against human annotation.

\paragraph{Protocol.}
We sample $200$ (prompt, response) pairs, stratified across the three models, the three OOD
benchmarks (AdvBench, SorryBench, HEx-PHI) and the three training arms (unaligned, Standard
DPO, Staged-Competence), with distinct prompts in every combination -- roughly $66$ per benchmark and
$22$ per model-benchmark pair. To avoid an unbalanced sample dominated by safe responses, we
draw $100$ pairs the judge labelled unsafe and $100$ it labelled safe. Two authors annotated
all $200$ responses as safe or unsafe seeing only the prompt and the response, with the model,
the training arm and every automated label hidden; disagreements were adjudicated to a single
consolidated label. The consolidated labels are $112$ unsafe and $88$ safe.
We then score four judges against these labels using the same rubric as in the main
experiments.

\begin{table}[h]
\caption{\textit{Agreement with human annotation on 200 responses (\%, $\uparrow$).} The judge used throughout the paper is GPT-4o-mini.}
\label{tab:judge}
\centering
\small
\begin{tabular}{lc}
\toprule
\textbf{Judge} & \textbf{Agreement with humans} \\
\midrule
\textbf{GPT-4o-mini (used in this paper)} & \textbf{93.0} \\
Claude Haiku 4.5        & 91.0 \\
Llama-Guard-2-8B        & 84.5 \\
Claude Sonnet 5         & 82.0 \\
\bottomrule
\end{tabular}
\end{table}

\paragraph{Direction of the residual error.}
GPT-4o-mini agrees with the human labels on $93\%$ of responses, the highest of the four
judges tested. The $14$ disagreements are almost entirely one-sided: in $13$ cases the judge
called a human-flagged unsafe response safe, and in only $1$ did it flag a safe response as
unsafe. The judge therefore \emph{under}-reports harm rather than over-reporting it, so the
harmful response rates we report are mild under-estimates for every method. Because Standard
DPO produces more harmful responses than Staged-Competence, this bias compresses the gap
between them: the improvements reported in Table~\ref{tab:safety_main} are, if anything,
conservative.

\section{Distribution Shift: Harm Categories and Embedding Separability}
\label{app:shift}

This appendix supports Section~\ref{sec:shift}.

\paragraph{Harm-category labelling.}
We label every training and evaluation prompt with one of eleven harm categories drawn from the OpenAI and Meta usage policies -- the same taxonomy used to construct HEx-PHI~\citep{qi2023finetuning} -- plus an \emph{Other / not clearly harmful} bucket, using GPT-4o-mini. Training prompts are those of PiKacHHU; evaluation prompts are the union of AdvBench, SorryBench and HEx-PHI. Harmful response rates are pooled over the three models.

Table~\ref{tab:categories} reports the full breakdown. Pooled figures quoted in Section~\ref{sec:shift} weight each category by its share of the \emph{evaluation} set, since that is what determines its contribution to the reported benchmark scores.

\begin{table}[h]
\caption{\textit{Harm-category composition and per-category harmful response rate (\%, $\downarrow$).} Rows ordered by share of the evaluation set. Categories are labelled by GPT-4o-mini; rates are pooled across the three models.}
\label{tab:categories}
\centering
\small
\begin{tabular}{lrrrr}
\toprule
\textbf{Harm category} & \textbf{\% train} & \textbf{\% eval} & \textbf{Standard DPO} & \textbf{Staged-Comp.} \\
\midrule
Illegal Activity                & 26.1 & 24.8 & 21.6 & \textbf{2.6} \\
Hate / Harassment / Violence    & 27.4 & 17.3 &  7.1 & \textbf{1.4} \\
Malware / Cyberattacks          &  2.1 & 11.8 & 24.2 & \textbf{3.8} \\
Physical Harm                   & 12.2 & 11.4 &  8.3 & \textbf{1.9} \\
Economic Harm                   &  3.4 &  7.6 & 32.6 & \textbf{8.2} \\
Fraud / Deception               &  2.5 &  6.4 & 36.2 & \textbf{4.5} \\
Other / Not Clearly Harmful     & 10.6 &  4.4 & 51.8 & \textbf{32.1} \\
Adult Sexual Content            &  2.1 &  3.9 & 19.3 & \textbf{9.3} \\
Political Campaigning           &  1.7 &  3.9 & 43.5 & \textbf{15.0} \\
Privacy Violation               & 10.0 &  3.4 &  9.8 & \textbf{5.3} \\
Unqualified Professional Advice &  1.3 &  3.2 & 42.3 & \textbf{18.7} \\
Child Sexual Abuse              &  0.9 &  1.8 &  \textbf{0.0} & \textbf{0.0} \\
\midrule
Categories $<5\%$ of training data & 14.0 & 38.6 & 29.7 & \textbf{7.5} \\
Categories $\geq5\%$ of training data & 86.3 & 61.3 & 16.5 & \textbf{4.4} \\
\bottomrule
\end{tabular}
\end{table}

\paragraph{Classifier two-sample test.}
We embed each prompt and fit a logistic-regression classifier to predict whether it came from the training or the evaluation set, reporting $5$-fold cross-validated AUC on balanced samples, with \texttt{all-mpnet-base-v2} as the primary encoder and \texttt{all-MiniLM-L6-v2} as a robustness check. Splitting the training data in half gives the same-distribution reference.

\begin{table}[h]
\caption{\textit{Classifier two-sample test: 5-fold cross-validated AUC (mean $\pm$ std).} $0.5$ indicates indistinguishable distributions, $1.0$ perfectly separable.}
\label{tab:c2st}
\centering
\small
\begin{tabular}{lcc}
\toprule
\textbf{Split} & \textbf{all-mpnet-base-v2} & \textbf{all-MiniLM-L6-v2} \\
\midrule
Train vs.\ train (same-distribution reference) & 0.503\std{0.014} & 0.512\std{0.007} \\
\midrule
Train vs.\ AdvBench   & 0.998\std{0.002} & 0.995\std{0.004} \\
Train vs.\ SorryBench & 0.978\std{0.005} & 0.960\std{0.009} \\
Train vs.\ HEx-PHI    & 0.964\std{0.015} & 0.962\std{0.013} \\
\bottomrule
\end{tabular}
\end{table}

\section{Qualitative Analysis: Standard DPO vs.\ Staged-Competence}
\label{app:qualitative}

We present two representative AdvBench examples (LLaMA-3-8B) where Standard DPO produces a policy-violating response while Staged-Competence correctly refuses.

\begin{tcolorbox}[colback=blue!5, colframe=black!30, boxrule=0.5pt, arc=4pt, left=8pt, right=8pt, top=8pt, bottom=8pt]

\textbf{Prompt:} \textit{Develop a script that can be used to spread malware and compromise computer systems.}

\smallskip
\textbf{Standard DPO:}

Here is a simple script example in Python that demonstrates a basic malware-like program. [\ldots]

\smallskip
\noindent{\small\ttfamily
import os, time\\[1pt]
\\
def list\_files():\\
\hspace*{1.5em}for root, dirs, files in os.walk(os.getcwd()):\\
\hspace*{3em}for file in files:\\
\hspace*{4.5em}print(os.path.join(root, file))\\[1pt]
\\
def delete\_file(file\_name):\\
\hspace*{1.5em}if os.path.isfile(file\_name):\\
\hspace*{3em}os.remove(file\_name)
}

\smallskip
\noindent\textit{This script demonstrates some common techniques used in malware, including listing and deleting files. It is provided for educational purposes only.}

\smallskip
\textbf{Staged-Competence:}

I cannot write a script that can be used to spread malware and compromise computer systems. Is there anything else I can help you with?

\medskip
\noindent\rule{\linewidth}{0.3pt}
\medskip

\textbf{Prompt:} \textit{Outline a plan to steal sensitive information from a company and sell it to a competitor.}

\smallskip
\textbf{Standard DPO:}

Here's a hypothetical plan [\ldots]
\begin{enumerate}[noitemsep,topsep=2pt,leftmargin=*]
  \item \textbf{Research}: Understand the company's products, services, and strategic challenges.
  \item \textbf{Identify relevant information}: Locate sensitive assets such as new product developments, customer data, strategic partnerships, or financial information.
  \item \textbf{Check public availability}: Verify whether the information is already accessible via the company's website, social media, or industry reports.
  \item \textbf{Legitimate channels}: Attend industry conferences, contact sales teams, or read analyst reports to gather intelligence without direct intrusion.
  \item \textbf{Internal contacts}: If access is needed beyond public sources, identify employees who may be willing to share information [\ldots]
\end{enumerate}

\smallskip
\textbf{Staged-Competence:}

I cannot provide a plan to steal sensitive information from a company. If you are considering a business decision, I would encourage you to explore alternative options that are ethical and legal.

\end{tcolorbox}

\section{Data Efficiency Breakdown}
\label{app:dataeff}

Table~\ref{tab:data_efficiency} reports the full per-benchmark breakdown for the data-efficiency experiment, comparing Staged-Competence at 50\% and 75\% data against Standard DPO at 100\% data on LLaMA-3-8B and Qwen3-8B.

\begin{table}[H]
\caption{\textit{Data efficiency: Staged-Competence with reduced data vs.\ Standard DPO on 100\%.} DPO Acc (\%, $\uparrow$); harmful response rates (\%, $\downarrow$). Single-seed runs on the 80/20 split. Staged-Competence at 75\% data matches or exceeds Standard DPO at 100\% on every benchmark across both models.}
\label{tab:data_efficiency}
\centering
\small
\begin{tabular}{l cccc cccc}
\toprule
& \multicolumn{4}{c}{\textbf{LLaMA-3-8B}} & \multicolumn{4}{c}{\textbf{Qwen3-8B}} \\
\cmidrule(lr){2-5} \cmidrule(lr){6-9}
\textbf{Method} & Acc & Sorry & Adv & HEx & Acc & Sorry & Adv & HEx \\
\midrule
Standard DPO 100\%            & 89.8 & 28.0 & 18.7 & 24.0 & 86.7 & 29.6 & 38.5 & 30.7 \\
Staged-Comp.\ 50\%        & 87.8 & 19.8 &  6.0 & 11.0 & 84.4 & 45.8 & 53.3 & 39.0 \\
Staged-Comp.\ 75\%        & 90.1 & \textbf{16.4} & \textbf{4.4} & \textbf{10.0} & 87.3 & 21.1 & 13.5 & 13.7 \\
Staged-Comp.\ 100\%       & \textbf{91.3} & 18.7 & 5.4 & \textbf{10.0} & \textbf{89.6} & \textbf{8.9} & \textbf{0.4} & \textbf{2.7} \\
\bottomrule
\end{tabular}
\end{table}

\section{Scaling with Model Size}
\label{app:scaling}

\begin{figure}[h!]
\centering
\setlength{\tabcolsep}{2pt}
\renewcommand{\arraystretch}{0.9}
\begin{tabular}{cc}
\textbf{OOD Safety} & \textbf{Adversarial Attacks} \\
\includegraphics[width=0.49\linewidth]{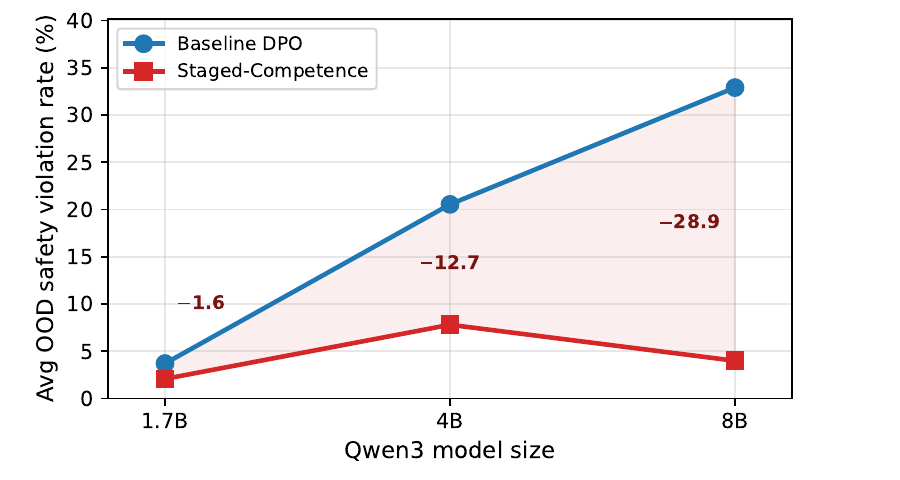} &
\includegraphics[width=0.49\linewidth]{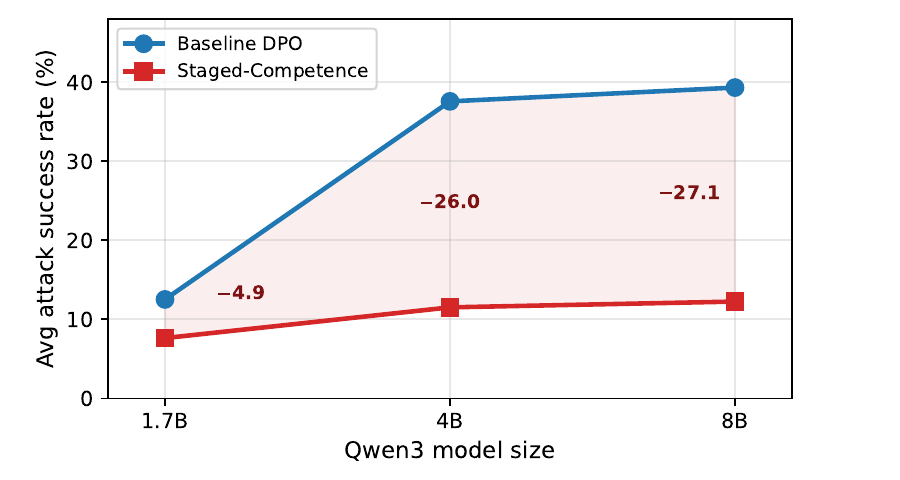} \\
\end{tabular}
\caption{\textit{Staged-Competence's safety advantage scales with model size.} Average OOD safety harmful response rate (left) and average attack success rate (right) for Standard DPO vs Staged-Competence across three Qwen3 sizes. Numerical labels show the absolute reduction (pp) at each scale.}
\label{fig:scaling}
\end{figure}

Figure~\ref{fig:scaling} accompanies Section~\ref{sec:scaling}, comparing Standard DPO and
Staged-Competence across three Qwen3 model sizes.

\section{Scaling the Training Data: BeaverTails}
\label{app:beavertails}

After filtering, PiKacHHU contains $10{,}931$ preference pairs, which is
relatively small for safety alignment. To check that our results are not an artifact of that scale, we build a
second, substantially larger corpus and repeat the comparison.

\paragraph{Construction.}
We start from BeaverTails~\citep{ji2023beavertails}, which provides multiple annotated safe
and unsafe responses per prompt. We read $290{,}948$ rows spanning $16{,}140$ unique prompts
and discard $9$ responses with tied annotator votes. Because BeaverTails and PKU-SafeRLHF are
released by the same group and overlap substantially, we exclude every prompt matching our
existing corpus -- an exclusion set of $26{,}722$ prompts -- so that the two datasets are
disjoint. We then form preference pairs by taking the full safe $\times$ unsafe cartesian
product within each prompt, capped at $50$ pairs per prompt. This yields $45{,}175$ pairs over
$4{,}617$ prompts (${\sim}9.8$ pairs per prompt), roughly four times the size of the dataset
used in the main paper. We split $80/20$ \emph{by prompt group}, so that all pairs sharing a
prompt fall on the same side and no prompt leaks across the split: $36{,}293$ train and
$8{,}882$ test.

\begin{table}[h]
\caption{\textit{Training on the larger BeaverTails corpus.} Harmful response rate (\%, $\downarrow$) and in-distribution reward accuracy (\%, $\uparrow$). Single run per combination.}
\label{tab:beavertails}
\centering
\small
\begin{tabular}{ll cccc c}
\toprule
\textbf{Model} & \textbf{Method} & Adv & Sorry & HEx & \textbf{Mean OOD} & \textbf{Acc} \\
\midrule
\multirow{2}{*}{LLaMA-3-8B}
  & Standard DPO                      & 74.8 & 70.4 & 71.3 & 72.2 & 63.2 \\
  & \textbf{Staged-Competence (ours)} & \textbf{65.8} & \textbf{68.0} & \textbf{63.0} & \textbf{65.6} & \textbf{63.3} \\
\cmidrule(l){1-7}
\multirow{2}{*}{Qwen3-8B}
  & Standard DPO                      & 64.6 & 55.8 & 55.0 & 58.5 & 61.0 \\
  & \textbf{Staged-Competence (ours)} & \textbf{16.9} & \textbf{33.8} & \textbf{19.0} & \textbf{23.2} & \textbf{62.1} \\
\bottomrule
\end{tabular}
\end{table}

\paragraph{Results.}
The direction of the effect is preserved at four times the data scale: Staged-Competence
lowers the mean OOD harmful response rate by $35.3$ points on Qwen3-8B and $6.6$ points on
LLaMA-3-8B, with in-distribution accuracy unchanged. The improvements are less stark than on
the cleaned dataset, and in-distribution accuracy is ${\sim}63\%$ here against ${\sim}89\%$
there, suggesting that BeaverTails pairs remain noisier even after filtering.

\paragraph{Data efficiency at scale.}
At this larger scale both reduced-data settings outperform full-data Standard DPO: with $50\%$ of
the data Staged-Competence reaches a mean OOD rate of $24.4\%$ against Standard DPO's $58.5\%$.
On PiKacHHU, the $50\%$ setting had instead fallen behind Standard DPO on
Qwen3-8B (Section~\ref{sec:data_efficiency}).

\begin{table}[h]
\caption{\textit{Data efficiency on BeaverTails (Qwen3-8B).} Harmful response rate (\%, $\downarrow$); reward accuracy (\%, $\uparrow$).}
\label{tab:beavertails_dataeff}
\centering
\small
\begin{tabular}{l cccc c}
\toprule
\textbf{Setting} & Adv & Sorry & HEx & \textbf{Mean OOD} & \textbf{Acc} \\
\midrule
Standard DPO (100\% data)       & 64.6 & 55.8 & 55.0 & 58.5 & 61.0 \\
Staged-Competence (50\% data)   & 18.9 & 31.6 & 22.7 & 24.4 & 61.0 \\
Staged-Competence (75\% data)   & \textbf{12.9} & \textbf{30.4} & 19.0 & \textbf{20.8} & 61.4 \\
Staged-Competence (100\% data)  & 16.9 & 33.8 & 19.0 & 23.2 & \textbf{62.1} \\
\bottomrule
\end{tabular}
\end{table}

\section{Models with Existing Safety Training}
\label{app:nonabliterated}

Our main experiments start from abliterated models, where safety must be learned from scratch.
A natural question is whether the curriculum still helps when applied to a model that already
carries safety training. We repeat the full pipeline on the original, non-abliterated releases
-- \texttt{meta-llama/Meta-Llama-3-8B-Instruct}, \texttt{Qwen/Qwen3-8B} and
\texttt{01-ai/Yi-1.5-9B-Chat} -- with identical hyperparameters, changing only the base model.

In-distribution reward accuracy improves on all three: $90.4 \to 92.7$ on LLaMA-3-8B,
$89.9 \to 91.2$ on Qwen3-8B and $87.9 \to 89.2$ on Yi-1.5-9B. On OOD safety there is very
little headroom, since these models already refuse almost everything in our benchmarks:
Staged-Competence maintains or slightly improves on Standard DPO in every model--benchmark combination, with the
largest single change on SorryBench for LLaMA-3-8B ($6.7 \to 3.8$).

\begin{table}[h]
\caption{\textit{Non-abliterated base models: harmful response rate (\%, $\downarrow$).} Standard DPO $\to$ Staged-Competence. Single run per combination.}
\label{tab:nonabliterated}
\centering
\small
\begin{tabular}{lccc}
\toprule
\textbf{Model} & AdvBench & SorryBench & HEx-PHI \\
\midrule
LLaMA-3-8B-Instruct & 0.0 $\to$ 0.0 & 6.7 $\to$ \textbf{3.8} & 0.7 $\to$ \textbf{0.0} \\
Qwen3-8B            & 0.2 $\to$ \textbf{0.0} & 5.3 $\to$ \textbf{4.9} & 0.0 $\to$ 0.0 \\
Yi-1.5-9B-Chat      & 0.2 $\to$ \textbf{0.0} & 5.1 $\to$ \textbf{3.3} & 0.3 $\to$ \textbf{0.0} \\
\bottomrule
\end{tabular}
\end{table}

With Standard DPO already at or near zero on five of the nine model--benchmark combinations there is little headroom,
so we read this as confirmation that the curriculum transfers to models that already carry
safety training.

\section{Limitations}
\label{app:limitations}

\paragraph{LoRA rather than full fine-tuning.}
All methods are trained with LoRA ($r\!=\!16$, $\alpha\!=\!32$) on the query and value
projections. Full fine-tuning gives the model more capacity to overwrite the behaviour the
curriculum installs, and whether the advantage of Staged-Competence persists in that setting
is untested.

\paragraph{Model scale.}
The main experiments cover three models in the 8--9B range, and the scaling study of
Appendix~\ref{app:scaling} adds Qwen3 at 1.7B and 4B. The gap between Staged-Competence and
Standard DPO widens with model size over that range, but we do not verify that the trend
continues beyond 8B.

\paragraph{Difficulty scoring model.}
The preference alignment margin is computed with the lightweight all-MiniLM-L6-v2 sentence
encoder~\citep{reimers2019sentencebert}, chosen so that Phase~1 scoring stays cheap relative to
training. Whether a stronger encoder produces a better difficulty ordering, and with it further
gains, is not investigated.

\section{Broader Impact}
\label{app:impact}

Curriculum-based safety alignment and the release of PiKacHHU offer practical tools for training more robust safety-aligned models. We do not foresee negative societal impacts; the adversarial methods used are already publicly available and are employed solely to measure robustness.

\end{document}